\documentclass[a4paper,fleqn]{cas-sc}
\usepackage{cite}
\usepackage{bbding}   
\usepackage{pifont}
\usepackage{float}
\usepackage{soul}
\usepackage{ulem} 
\usepackage{xcolor}
\usepackage{mdframed}
\usepackage[hypcap=false]{caption}
\usepackage{subfig,graphicx}
\usepackage[capitalize]{cleveref}
\crefname{section}{Sec.}{Secs.}
\Crefname{section}{Section}{Sections}
\Crefname{table}{Table}{Tables}
\crefname{table}{Tab.}{Tabs.}
\definecolor{lightgray}{rgb}{0.8, 0.8, 0.8}
\newcommand{\cmark}{\ding{51}\xspace}%
\newcommand{\xmarkg}{\textcolor{lightgray}{\ding{55}}\xspace}%

\def\tsc#1{\csdef{#1}{\textsc{\lowercase{#1}}\xspace}}
\tsc{WGM}
\tsc{QE}

\begin{document}
\let\WriteBookmarks\relax
\def\floatpagepagefraction{1}
\def\textpagefraction{.001}
\let\printorcid\relax
\shorttitle{}  
\shortauthors{Y. Bai et~al.}

\title[mode = title]{Dual-path Frequency Discriminators for Few-shot Anomaly Detection}

\nonumnote{
$^*$ Equal contribution. $\dagger$ Corresponding author. 
Email address: yhbai@zju.edu.cn (Yuhu Bai), 186368@zju.edu.cn (Jiangning Zhang), Guanzhong Tian (gztian@zju.edu.cn)
}
\author[1, 2]{\textcolor{black}{Yuhu Bai} $^{*}$}

\author[3,4]{\textcolor{black}{Jiangning Zhang} $^{*}$}

\author[5]{\textcolor{black}{Zhaofeng Chen}}

\author[1,2]{\textcolor{black}{Yuhang Dong}}

\author[6]{\textcolor{black}{Yunkang Cao}}

\author[1]{\textcolor{black}{Guanzhong Tian} $^{\dagger}$}

\address[1]{Ningbo Innovation Center, Zhejiang University, Ningbo 315100, China}
\address[2]{Polytechnic Institute, Zhejiang University, Hangzhou 310015, China}
\address[3]{College of Control Science and Engineering, Zhejiang University, Hangzhou 310027, China}
\address[4]{Youtu Lab, Tencent, Shanghai 200233, China}
\address[5]{China Tower (Hangzhou) Science and Technology Innovation Center, Hangzhou 310020, China}
\address[6]{School of Mechanical Science and Engineering, Huazhong University of
Science and Technology, Wuhan 430074, China}

\begin{abstract}
Few-shot anomaly detection (FSAD) plays a crucial role in industrial manufacturing. However, existing FSAD methods encounter difficulties leveraging a limited number of normal samples, frequently failing to detect and locate inconspicuous anomalies in the spatial domain. We have further discovered that these subtle anomalies would be more noticeable in the frequency domain. In this paper, we propose a Dual-Path Frequency Discriminators (DFD) network from a frequency perspective to tackle these issues. 
The original spatial images are transformed into multi-frequency images, making them more conducive to the tailored discriminators in detecting anomalies. Additionally, the discriminators learn a joint representation with forms of pseudo-anomalies. 
Extensive experiments conducted on MVTec AD and VisA benchmarks demonstrate that our DFD surpasses current state-of-the-art methods. The code is available at \url{https://github.com/yuhbai/DFD}.
\end{abstract}


	
		

\maketitle

\section{Introduction}
\label{sec:introduction}
Industrial images anomaly detection involves identifying anomalous samples in addition to precisely locating anomalies ~\cite{liu2024deep,cao2024survey, REB, MSTAD}. However, anomalies in industrial images encompass a wide range of types and occur infrequently. The acquisition of anomalous samples and the creation of labels for anomalous images present significant challenges in real-world applications. As a result, the majority of research is concentrated on unsupervised anomaly detection and localization. Currently, embedding-based~~\cite{bergmann2020uninformed, defard2021padim, roth2022towards, LMKAD, OMAC, jiang2023masked} methods and reconstruction-based~~\cite{bergmann2018improving,zavrtanik2021draem,you2022unified,wyatt2022anoddpm,liang2023omni} methods are the predominant methodologies for addressing this challenging issue. 

\begin{figure}[t]
    \centering
    \includegraphics[width=1\linewidth,trim=0 300 0 0,clip]{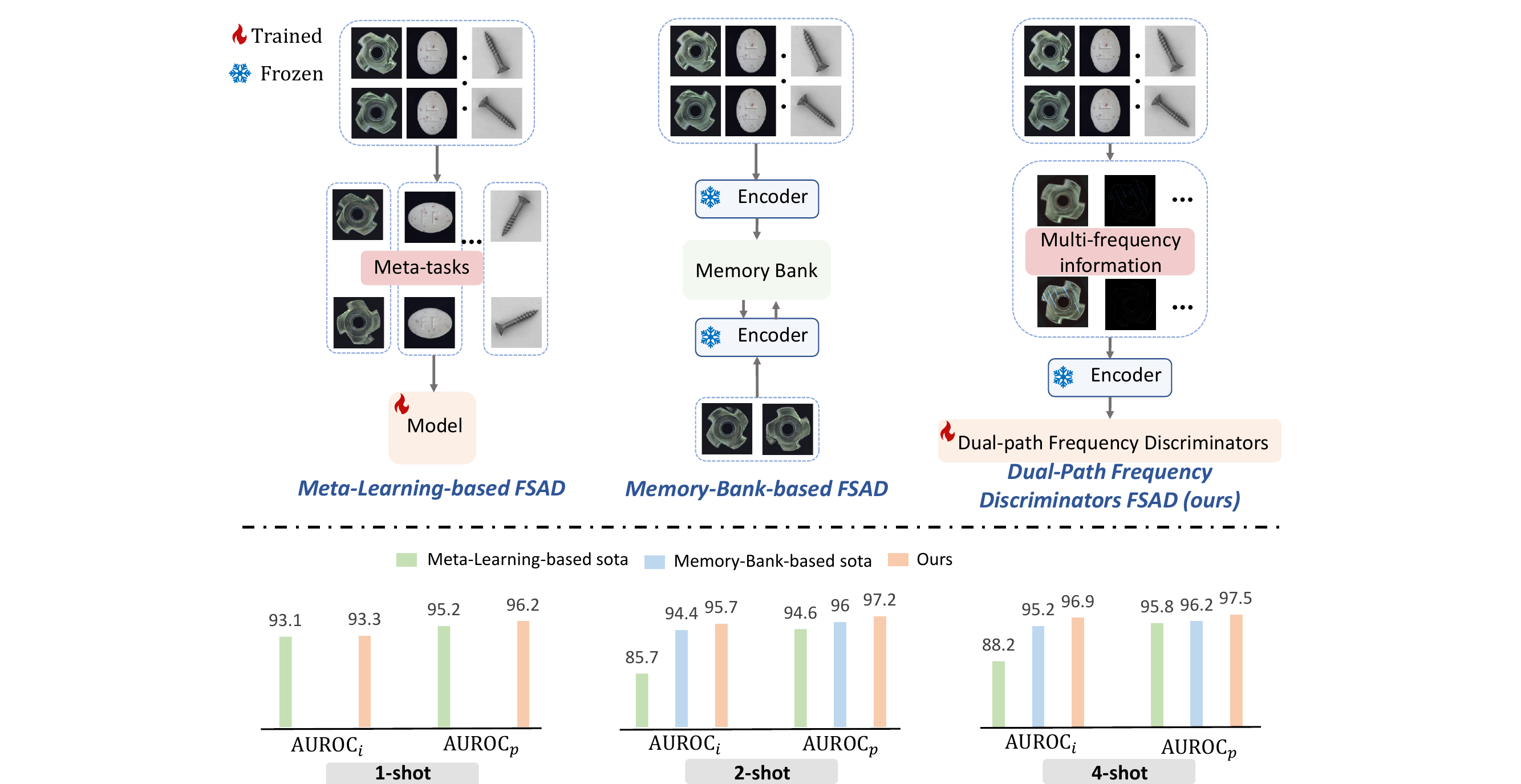}
    \caption{
    The comparison between DFD and sota methods. The top figure is previous FSAD framework v.s. ours. Comparison with meta-learning-based FSAD, our model is simple and stability. Comparison with memory-bank-based FSAD, our method needs no extra memory to restore features.
    The bottle figure is comparison with previous sota performance on MVTec AD dataset for 2-/4-shot setting.  
    }
    \label{fig:comparison}
\end{figure}

Considering the significant resources required to collect a substantial number of samples and the inherent similarities among industrial images within the same category, there is a growing interest in FSAD~~\cite{wu2021learning,huang2022registration,xie2023pushing,santos2023optimizing,fang2023fastrecon, maeday}. FSAD seeks to achieve performance comparable to full-shot anomaly detection methods with only a limited number of source images (less than 8). 
As illustrated in \cref{fig:comparison}, current FSAD methods can be broadly categorized into meta-learning-based methods and memory-bank-based methods.
Meta-learning-based FSAD, such as RegAD~\cite{huang2022registration} and Metaformer~\cite{wu2021learning}, leverage meta-learning strategy to deal with the problem of insufficient training samples. Memory-bank-based~\cite{xie2023pushing,santos2023optimizing,fang2023fastrecon} methods, on the other hand, attempt to employ feature matching for FSAD. 
However, these methods have some limitations: 
(1) They have not fully utilized the limited number of training images available; 
(2) Subtle anomalies are less noticeable in the spatial domain;
(3) Memory-bank-based methods do not effectively transfer the feature distribution from the images used in pre-trained models to industrial images. They also require additional memory bank to store features;
(4) Meta-learning-based methods have disadvantages of instability during training and enormous computational cost.

In order to solve the aforementioned challenges, we propose our Dual-path Frequency Discriminators (DFD) for FSAD. 
First, we broaden the dataset through straightforward data augmentation to maximize the utility of the limited number of samples.
Second, rather than relying solely on spatial information, we advocate for decoupling images into different frequency components. High-frequency components capture fine texture features within the image, while low-frequency components are associated with semantic information. 

\begin{wrapfigure}{r}{0.5\textwidth}
    \centering
    \vspace{-0.3cm}
    \includegraphics[width=\linewidth,trim=0 120 0 120,clip]{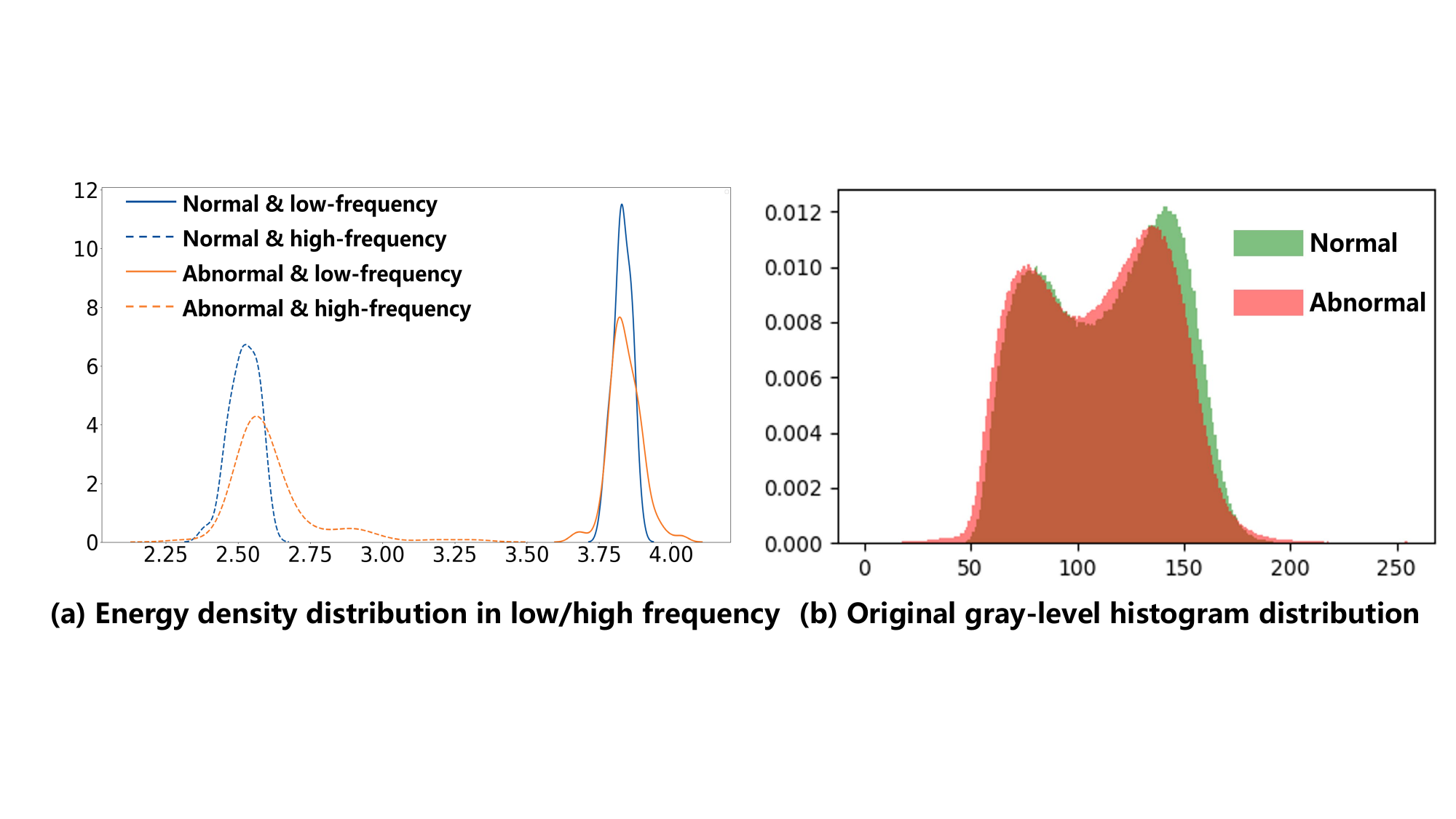}
    \caption{\textbf{Energy density distribution and gray-level histogram distribution of tile category.} (a) \textbf{Energy density distribution} in low-/high- frequency of tile category, showing that normal/abnormal images obviously differ in frequency distribution. (b) \textbf{Original gray-level histogram distribution} of tile category, showing that normal/abnormal images are hard to distinguish in spatial domain.}
    \vspace{-0.3cm}
\label{fig:frequency}
\end{wrapfigure}
Different types of anomalies manifest as alterations in various frequency bands, making subtle and imperceptible anomalies in the spatial domain more noticeable in the frequency domain. 
We further tally the information from the MVTev AD dataset in the spatial and frequency domain. \cref{fig:frequency} (b) shows that the spatial domain gray-level histogram cannot distinguish normal and abnormal images. 
However, \cref{fig:frequency} (a) reveals that the normal and abnormal images in the tile category exhibit different energy distributions at low and high frequencies (to obtain the energy density distribution, a two-dimensional Fourier transform\cite{fft} is performed on the image, resulting in a complex matrix. The Euclidean distance of matrix elements at coordinates $(x,y)$ from the center is indicative of the frequency value, with the modulus of the complex number representing energy. The energy distribution curve is plotted with the abscissa representing the distance from the point to the center (frequency) and the ordinate representing the amplitude value (energy)). 
Third, we suggest using a feature adaptor to alleviate domain bias and pull normal features together while push the anomaly features apart from normal features.
Finally, given that abnormal and normal images exhibit disparate feature distributions, it is feasible to determine the abnormality directly through the deployment of simple dual-path frequency discriminators without the need for an additional memory bank in the feature space. Training a discriminative network exclusively with normal images can lead to over-fitting, and the discriminative network cannot be optimized due to the absence of positive samples (i.e., anomalous samples). Therefore, we synthesize anomalies at both image-level and feature-level to facilitate the dual-path discriminators to consciously distinguish between normal and abnormal features. Although synthetic anomalies are not identical to real-world anomalies, they only need to differ from the normal feature distribution to effectively train discriminators capable of recognizing anomalies. 
Our main contributions are summarized as follows:
\begin{itemize}
    \item We approach anomaly detection as a classification problem from a frequency perspective. We present a novel and robust framework that effectively leverages a limited number of normal source images.
    \item A pseudo-anomaly generation strategy is designed to generate different forms of anomalies at image-level and feature-level. We propose multi-frequency information construction module and fine-grained feature construction module to obtain different frequency adapted features, which are subsequently fed into the Dual-path feature discrimination module. This module estimates abnormality in the latent space, enhancing the overall anomaly detection capability.
    \item We conduct extensive experiments on MVTec and VisA benchmarks, showing that our model outperforms previous FSAD methods. Specifically, our DFD exceeds previous state-of-the-art~\cite{jeong2023winclip}, improving MVTec AD by 1.3\% and 1.2\% at image-level AUROC and pixel-level AUROC under 2-shot scenarios.
\end{itemize}

\section{Related Work}
\subsection{Frequency decoupling}
Images are typically represented in the spatial domain, where the intensity value of each pixel represents the brightness or color of the image. 
The frequency domain represents the frequency and amplitude of various patterns and fluctuations within the image.
The frequency decoupling primarily involves the Fourier Transform and related concepts.
Specifically, the Fourier Transform \cite{fft} decomposes an image into a series of sinusoidal components, representing it in the frequency domain by their amplitudes and phases.
Consequently, the 2D Discrete Fourier Transform (DFT) for an image $f(x,y)$ of size $M \times N$ is given:
\begin{equation}
\centering
F(u, v)=\sum_{x=0}^{M-1} \sum_{y=0}^{N-1} f(x, y) e^{-j 2 \pi\left(\frac{u x}{M}+\frac{v y}{N}\right)},
\end{equation}
where $F(u,v)$ is the frequency representation at coordinates $(u,v)$, $j$ is the imaginary unit. In the Fourier space, the representation can be described by both amplitude $\mathcal{A}(u, v)$ and phase $\mathcal{P}(u, v)$:
\begin{equation}
\begin{aligned}
    \mathcal{A}(u, v) & =\left[R^2(x,y)(u, v)+I^2(x,y)(u, v)\right]^{1 / 2} \\
    \mathcal{P}(u, v) & =\arctan \left[\frac{I(x,y)(u, v)}{R(x,y)(u, v)}\right],
\end{aligned}
\end{equation}
where $R(x,y)$ and $I(x,y)$ denote the real and imaginary part of the image $f(x,y)$. 
In image processing, the amplitude $\mathcal{A}(u, v)$ typically indicates the prominence of different frequency fluctuations within an image. Meanwhile, the phase $\mathcal{P}(u, v)$ provides crucial information about each frequency component's phase, representing the relative shift of the waveform with respect to a reference point.

\subsection{Few-shot Learning}
Few-shot learning (FSL) pertains to the identification and classification of novel data utilizing an exceedingly limited quantity of training data.
FSL methods can be primarily categorised into model fine-tuning, transfer learning, and data augmentation. 
Fine-tuning methods~\cite{nakamura2019revisiting,liu2022few} typically involve pre-training models on large-scale datasets and then fine-tuning the fully connected layers of the model on a target few-shot dataset to obtain the fine-tuned model.
Transfer learning methods~\cite{wang2016learning,jang2019learning,finn2017model,xing2002distance} efficiently transfer the acquired knowledge to a new domain. 
Data augmentation methods~\cite{benaim2018one,aksu2021n,boudiaf2020information} perform data expansion or feature enhancement on the original few-shot dataset.

\subsection{Industrial Anomaly Detection}  
Existing anomaly detection methods are conventionally classified into three distinct categories.
\textbf{\textit{1)} Reconstruction-based methods}~\cite{bergmann2018improving,zavrtanik2021draem,wyatt2022anoddpm,diad,vitad,yao2023scalable,SSMCTB} posit that anomalous regions cannot be reconstructed using encoder-decoder architecture. Anomaly detection is performed by measuring the reconstruction errors of test samples. Autoencoder (AE), generative adversarial networks (GANs), Transformer, and diffusion model are utilized to reconstruct normal images. IMRN~\cite{WU2024111594} leverages a horizontal-vertical latent space to enhance reconstruction quality and module interactivity. OCR-GAN ~\cite{liang2023omni} employs omni-frequency representations in the reconstruction-based methods. PNPT ~\cite{yao2024prior} combines normal images as prompt to alleviate "identical mapping" during reconstruction.
\textbf{\textit{2)} Synthesizing-based methods} synthesize anomalies on normal samples~\cite{li2021cutpaste,zavrtanik2021draem,anomalydiffusion, li2022eid}. CutPaste~\cite{li2021cutpaste} constructs anomalous images by cutting out portions of anomaly-free images and pasting them onto other locations. The anomalous images in DRAEM~\cite{zavrtanik2021draem} are generated using Perlin noise. A reconstructive sub-network is trained to reconstruct the generated anomalous images into normal images, followed by inputting both the reconstructed images and the anomalous images into a segmentation network to predict the anomalous regions. 
\textbf{\textit{3)} Embedding-based methods} ~\cite{gudovskiy2022cflow, roth2022towards, liu2023simplenet, cdo, cao2024bias, cao2022informative,CHI2024112225}typically use a pre-trained network to extract features from normal samples. 
These methods differentiate normal and anomalous features by analyzing extracted shallow features. 
Mapping the feature distribution obtained from pre-trained models to a multivariate Gaussian distribution is also widely used. Several works~\cite{gudovskiy2022cflow, yao2023dual} employ normalization flow to construct a reversible mapping from original feature distribution to normal feature distribution. 
PatchCore~\cite{roth2022towards} proposes an efficient algorithm for striking a balance retaining a maximum amount of nominal patch features and minimal runtime through coreset subsampling. SimpleNet~\cite{liu2023simplenet} uses a simple discriminator composed of a 2-layer multi-layer perceptron(MLP) to detect and locate anomalies. 

\subsection{Few-shot Anomaly Detection}  
Recently, researchers have been increasingly concerned about \textbf{FSAD}. The objective of FSAD is to establish competitiveness in comparison to prevailing full-shot anomaly detection methods.  
Some works~\cite{wu2021learning,huang2022registration} leverage the meta-learning paradigm for training, which requires a substantial amount of base data to construct meta-tasks. RegAD employs a Siamese Neural Network framework, augmented with a Spatial Transformer Network (STN) to facilitate precise feature registration.
While others~\cite{xie2023pushing,santos2023optimizing} optimize PatchCore~\cite{roth2022towards} for few-shot setting. 
With the success of vision-language models, recent methods have integrated these models into AD. FOADS~\cite{FOADS} utilizes a framework based on Neural Gas (NG) network to extract feature embedding.
WinCLIP~\cite{jeong2023winclip} proposes a window-based CLIP framework for FSAD via fine-grained textual definitions and normal reference samples for feature matching. However, these optimizations often suffer from feature bias. 

In this work, we introduce a DFD framework tailored for few-shot anomaly detection from a frequency perspective. This method meticulously developed distinct modules to systematically address the aforementioned challenges.
\section{Method}
\begin{figure*}[tb]
  \centering
    \includegraphics[width=\textwidth,trim=0 173 0 0,clip]{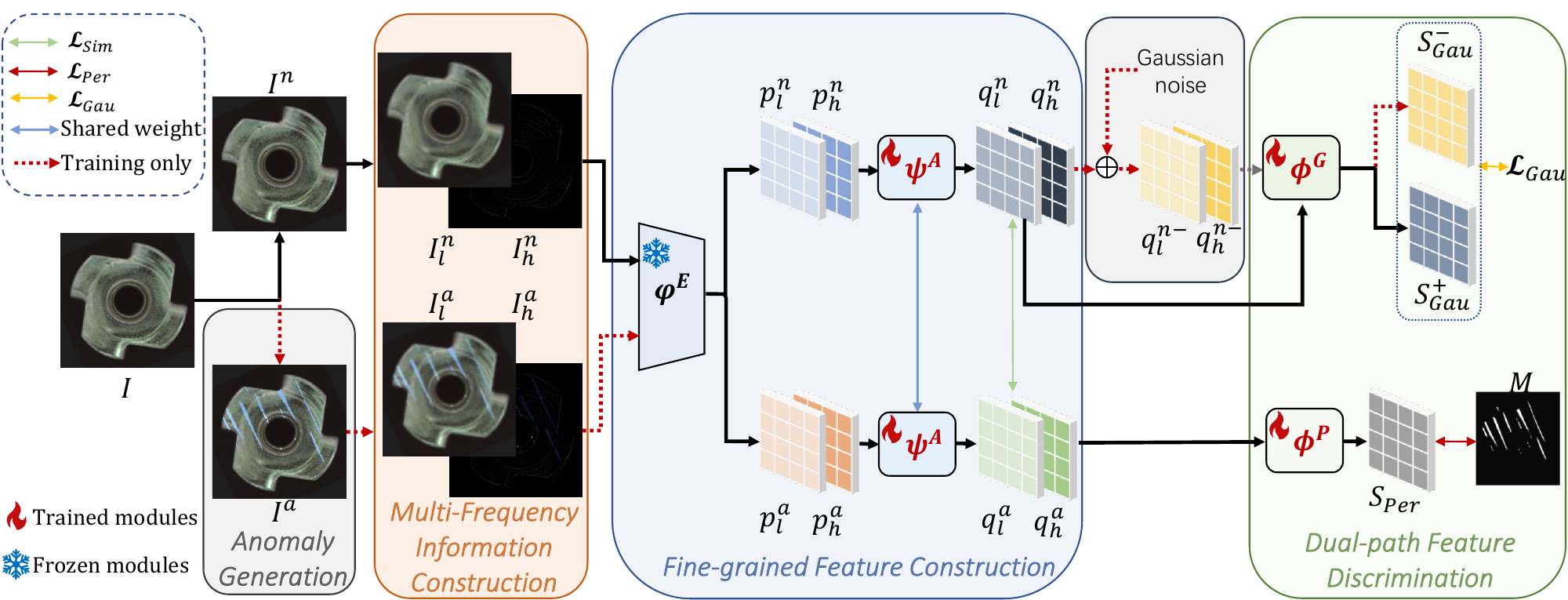}
  \caption{
  \textbf{Overview of proposed DFD framework}, which mainly consists of: \textbf{1) Anomaly Generation} module in \cref{anomaly generation}; \textbf{2) Multi-Frequency Information Construction} module in \cref{multi-frequency}; \textbf{3) Fine-grained Feature Construction} module in \cref{Fine-grained Feature Construction}; and \textbf{4) Dual-path Feature Discrimination} module in \cref{dual-path}. Input image $I$ is used to generate normal image $I^n$ and abnormal image $I^a$, which are then decoupled into different frequency components by Multi-Frequency Information Construction module, obtaining $I_l^n$/$I_h^n$ and $I_l^a$/$I_h^a$. Fine-grained Feature Construction takes above components as inputs that go through a pre-trained feature extractor $\varphi^E$ to extract local feature $p_l^n$/$p_h^n$ and $p_l^a$/$p_h^a$. Subsequent feature adaptor $\psi^A$ further transforms local feature to adapted feature $q_l^n$/$q_h^n$ and $q_l^a$/$q_h^a$. Gaussian noise is added to normal features $q_l^n$/$q_h^n$ to get pseudo-anomalous features $q_l^{n-}$/$q_h^{n-}$. Dual-path Feature Discrimination module contains Gaussian Discriminator $\phi^G$ estimating anomalies $S_{Gau}^-$ and $S_{Gau}^+$ for $q_l^{n-}$/$q_h^{n-}$ and $q_l^n$/$q_h^n$, and Perlin Discriminator $\phi^P$ estimating anomalies $S_{Per}$ for $p_l^{a}$/$p_h^{a}$.}
  \label{fig:overview}
\end{figure*}
The proposed DFD contains 4 parts: anomaly generation (\cref{anomaly generation}), multi-frequency information construction (\cref{multi-frequency}), fine-grained feature construction (\cref{Fine-grained Feature Construction}), and dual-path feature discrimination (\cref{dual-path}). By leveraging frequency information instead of spatial information, the dual-path discriminators network can more effectively identify anomalies. The discriminators are capable of learning joint representation from both normal images and pseudo-anomalies. The overview of our method is illustrated in \cref{fig:overview}.

\subsection{Anomaly Generation}
\label{anomaly generation}
\begin{figure}
    \centering
    \includegraphics[width=0.6\linewidth, trim=185 290 185 0,clip]{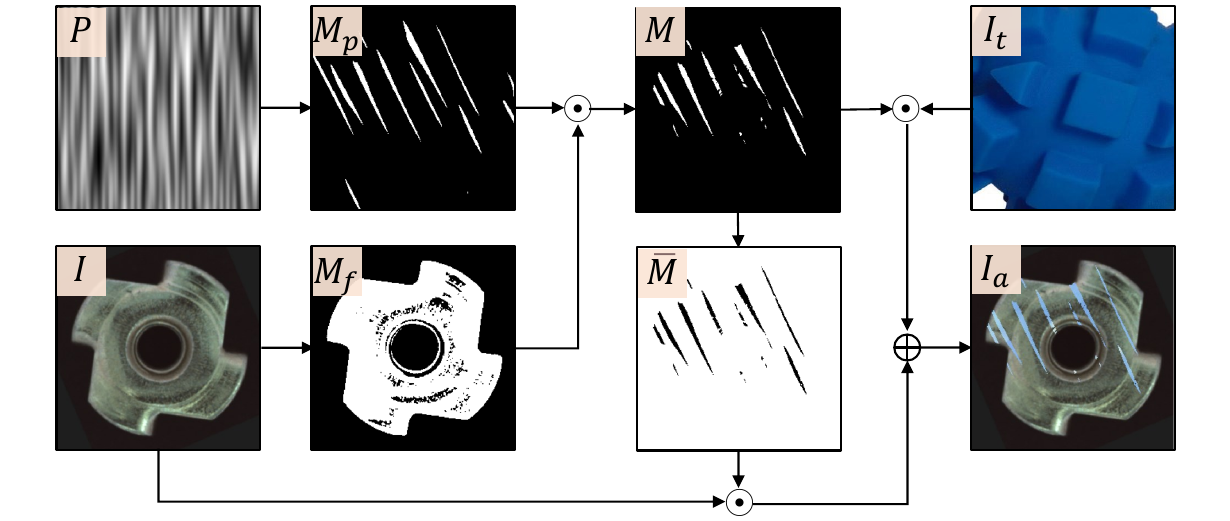}
    \captionof{figure}{\textbf{Image-level anomaly generation strategy}. The mask $M$ is obtained by performing element-wise product on $M_p$ and $M_f$ which are generated from random Perlin noise and source normal image. The pseudo-anomalous image is generated from $I$/$I_t$ according to $M$.}
    \label{fig:anomaly imag}
\end{figure}

Anomaly detection assumes that the feature distribution of anomaly-free samples follows a normal distribution. Intuitively, we can construct image-level pseudo-anomalies on normal images. Furthermore, to create feature-level pseudo-anomalies that deviate from the normal distribution, we introduce noise to the features of normal samples at the feature-level. This approach allows us to generate various forms of anomalies from different perspectives during training. The anomaly generation strategy is detailed below.
\hspace*{\fill} \\
\noindent\textbf{Image-level anomaly generation}.
As shown in \cref{fig:anomaly imag}, pseudo-anomalous images are generated based on normal images following DRAEM ~\cite{zavrtanik2021draem}. Initially, an original normal image $I \in \mathbb{R}^{H \times W \times 3}$ undergoes binarization to yield a foreground image mask $M_f$. Subsequently, a 2-dimensional Perlin noise \textit{P} is randomly generated and subjected to threshold-based binarization to generate a noise mask $M_p$. To ensure pseudo-anomalies only appear on the foreground image, an anomaly mask $M$ is generated by performing an element-wise product on $M_f$ and $M_p$ . 

A texture image $I_t$ is then masked with an anomaly mask $M$. To achieve a balanced fusion of the original normal image and the noise image, a transparency factor $\beta$ is introduced, facilitating a closer resemblance of the generated anomaly patterns to real anomalies. Therefore, the generated pseudo-anomalous image $I_a$ is defined as:
\begin{equation}
\begin{gathered}
    I_a = \bar{M} \odot I+(1-\beta)\left(M \odot I\right)+\beta\left(M \odot I_t\right),\\
    M = M_f \odot M_p,
    \label{eq:anomaly image}
\end{gathered}
\end{equation}
where $\bar{M}$ is the inverse of $M$, $\odot$ is Hadamard Product.

\hspace*{\fill} \\
\noindent
\textbf{Feature-level anomaly generation}.
For the feature-level pseudo-anomaly generation, a Gaussian noise $\epsilon$ is randomly sampled from i.i.d Gaussian distribution $\mathcal{N}(\mu,\sigma^2)$, which is added to normal features $q_l^{n}/q_h^{n} \in \mathbb{R}^{h \times w \times C}$ in \cref{Fine-grained Feature Construction} to obtain pseudo-anomalous features $q_l^{n-}/q_h^{n-}$ in different frequency components:
\begin{equation}
    q_l^{n-} = q_l^{n} + \epsilon, q_h^{n-} = q_h^{n} + \epsilon.
    \label{eq:feature-level anomaly generation}
\end{equation}

\subsection{Multi-Frequency Information Construction}
\label{multi-frequency}
Various frequency components encompass distinct information, and different anomalies result in altered information within specific frequency bands. As shown in \cref{fig:frequency}, normal and abnormal samples have different energy distributions at low and frequencies. Thus, unlike the spatial domain, the frequency domain provides a novel perspective for anomaly detection.    

Given an image $I^{'}$, we convolve it with a Gaussian kernel and then remove all even rows and even columns to obtain intermediate image $I_{inter}$. We denote the above process as $\mathrm{Down}$. Next, we perform operation $\mathrm{Up}$ by expanding $I_{inter}$ to twice its original size in each dimension, filling new rows and columns (even rows and columns) with zeros. Subsequently, convolution is performed to approximate missing pixels with a Gaussian kernel. The low-frequency image $I_l$ is acquired:
\begin{equation}
    I_l = \mathrm{Up}(\mathrm{Down}(I^{'})).
\end{equation}
To recover the missing information, denoted as high-frequency image $I_h$, we compute the difference between the original image $I^{'}$ and low-frequency image $I_l$, which is represented as follows:
\begin{equation}
    I_h = I^{'} - I_l.
\end{equation}
We carry out above operations for both normal and pseudo-anomalous images, getting their multi-frequency information $I_l^n$/$I_h^n$ and $I_l^a$/$I_h^a$.

\subsection{Fine-grained Feature Construction}
\label{Fine-grained Feature Construction}
The fine-grained feature construction module comprises a feature extractor $\varphi^E$ and a feature adaptor $\psi^A$, which is anticipated to obtain adapted features for industrial images.

Following PatchCore~\cite{roth2022towards}, we use a pre-trained WideResnet-50~\cite{zagoruyko2016wide} as the feature extractor $\varphi^E$ to extract local features from multi-frequency information $I_l^n$/$I_h^n$ and $I_l^a$/$I_h^a$. However, since the pre-training dataset exhibits different distributions from industrial images, we incorporate a feature adaptor $\psi^A$ to mitigate the domain bias. Besides, we aim to make the boundary between abnormal and normal features more distinct both before and after they pass through the feature adaptor. The adaptor consists of a single linear layer without any activation function. Taking the low-frequency component of a normal image $I_l^n$ as an example, the adapted feature is defined as follows:
\begin{equation}
    p_l^n = \varphi^E(I_l^n), q_l^n = \psi^A(p_l^n),\\
\end{equation}
where $p_l^n$ is the local features. Through the aforementioned process, we get the adapted feature $q_l^n$/$q_h^n, q_l^a$/$q_h^a \in \mathbb{R}^{h \times w \times C}$.

\subsection{Dual-path Feature Discrimination}
\label{dual-path}
The feature distributions of the normal and abnormal samples exhibit differences, with the adapted features providing spatial information. By formulating anomaly detection as a feature space classification problem, we can effectively assess the abnormality of the adapted features. In this section, we present a dual-path feature discrimination module, comprising a Gaussian discriminator $\phi^G$ and a Perlin discriminator $\phi^P$, to identify pseudo-anomalies generated at both the feature-level and image-level.

\noindent
\textbf{Gaussian Discriminator}.  
In this branch, the normal adapted features $q_l^n$/$q_h^n\in \mathbb{R}^{h \times w \times C}$ and pseudo-anomalous features $q_l^{n-}$/$q_h^{n-}\in \mathbb{R}^{h \times w \times C}$ are forwarded to Gaussian Discriminator $\phi^G$ to estimate the abnormality at each position $(h, w)$. The output $\phi^G(q) \in \mathbb{R}^{h \times w}$ of Gaussian Discriminator is positive for normal features while negative for pseudo-anomalous features. The Gaussian discriminator $\phi^G$ is constructed using a 2-layer multi-layer perceptron (MLP) structure.

\noindent
\textbf{Perlin Discriminator}. 
Vision Transformer (ViT) leverages the self-attention mechanism to capture global long-term dependencies, enabling the model to understand contextual relationships across the entire image. Moreover, ViT is able to recognize intricate patterns and details~~\cite{he2022transfg, khan2022transformers}. These attributes are beneficial for comprehending anomalies in industrial scenarios.
Similar to the Gaussian Discriminator $\phi^G$, the output of the Perlin Discriminator $\phi^P(q) \in \mathbb{R}^{h \times w}$ is expected to be positive for normal features while negative for abnormal features at each position $(h, w)$. 
We construct the Perlin Discriminator $\phi^P$ by combining a single-layer MLP and a single-layer ViT.

\subsection{Training Objectives}
We propose three losses for training DFD in \cref{fig:overview}.

\noindent
\textbf{Similarity loss.}  
In order to push the anomalous features apart from normal features and pull the normal features together, the similarity loss $\mathcal{L}_{Sim}$ is utilized between pseudo-anomalous images and normal images at corresponding positions:
\begin{equation}
\left\{
\begin{aligned}
    \mathcal{L}_{l_{Sim}} &= 1 - \cos(M' \odot q_l^a, M' \odot q_l^n), \\
    \mathcal{L}_{h_{Sim}} &= 1 - \cos(M'\odot q_h^a, M' \odot q_h^n), \\
    \mathcal{L}_{Sim} &= \mathcal{L}_{l_{Sim}} + \mathcal{L}_{h_{Sim}},
\end{aligned}
\right.
\end{equation}
where $M' \in \mathbb{R}^{h \times w}$ is yielded by applying max pooling to $M \in \mathbb{R}^{H \times W}$. During training, we encourage feature adaptor to separate normal features from anomaly features, while ensuring normal features remain compact. Strong differences between the pseudo-anomalous and normal images are ensured by optimizing the similarity loss $\mathcal{L}_{Sim}$.

\noindent
\textbf{Gaussian loss.}
Gaussian loss penalizes negative scores for normal features and positive for pseudo-anomalous features following.
We use truncated $l_1$ loss as Gaussian loss:
\begin{equation}
\left\{
\begin{aligned}
    \mathcal{L}_{l_{Gau}}&= \max\{0,\theta - \phi^G(q_l^n)\} + \max\{0,\theta + \phi^G(q_l^{n-})\}, \\
    \mathcal{L}_{h_{Gau}}&= \max\{0,\theta - \phi^G(q_h^n)\} + \max\{0,\theta + \phi^G(q_h^{n-})\}, \\
    \mathcal{L}_{Gau} &= \mathcal{L}_{l_{Gau}} + \mathcal{L}_{h_{Gau}},
\end{aligned}
\right.
\label{gaussian loss}
\end{equation}
where $\theta$ is set to 0.8 by default preventing over-fitting.

\noindent
\textbf{Perlin loss.}
First, truncated $l_1$ loss is employed to ensure that Perlin Discriminator $\phi^P$ can locate the generated pseudo-anomalous regions:
\begin{equation}
\begin{aligned}
    \mathcal{L}_{l_{pix}}=&\max \{0,\theta - \phi^P(q_l^a) \odot (1-M') \} + \\
    &\max\{0,\theta + \phi^P(q_l^a)\odot M'\}.
\end{aligned}
\label{eq:low-frequency pixel loss}
\end{equation}
The high-frequency loss $\mathcal{L}_{h_{pix}}$ is similar to \cref{eq:low-frequency pixel loss}. Consequently, the pixel loss is defined as:
\begin{equation}
    \mathcal{L}_{pix} = \mathcal{L}_{l_{pix}} + \mathcal{L}_{h_{pix}}.
    \label{pixel loss}
\end{equation}
What's more, the maximum value of the output of $\phi^P$ is taken to estimate abnormality for the image:
\begin{equation}\
\left\{
\begin{aligned}
    \mathcal{L}_{l_{cls}} &= ||\tau - {\max}\{\mathrm{Sigmoid}(-\phi(q_l^a))\}||^2,\\
    \mathcal{L}_{h_{cls}} &= ||\tau - {\max}\{\mathrm{Sigmoid}(-\phi(q_h^a))\}||^2,\\
    \mathcal{L}_{cls} & =\mathcal{L}_{l_{cls}}+\mathcal{L}_{h_{cls}},
\end{aligned}
\right.
\end{equation}
where $\tau$ is the ground truth of the image abnormality. The overall Perlin loss $\mathcal{L}_{\text {Per }}$ is defined as :
\begin{equation}
    \mathcal{L}_{\text {Per}} = \frac{1}{2}(\mathcal{L}_{\text {pix}} + \mathcal{L}_{\text {cls}}).
\end{equation}

In summary, the total loss is defined as:
\begin{equation}
    \mathcal{L} = \mathcal{L}_{\text {Gau}} + \lambda_{Per} \mathcal{L}_{\text {Per}} + \lambda_{Sim} \mathcal{L}_{\text {Sim}}.
    \label{eq:total loss}
\end{equation}

\subsection{Inference}
As depicted in \cref{fig:overview}, the process of generating anomalies at image-level and feature-level is discarded during inference. For a test image $I_{test}\in \mathbb{R}^{H \times W \times 3}$, we obtain its low-/high-frequency adapted features $q^l/q^h \in \mathbb{R}^{h \times w \times C}$. Gaussian Discriminator $\phi^G$ and Perlin Discriminator $\phi^P$ calculate the anomaly scores $S_{Gau}, S_{Per} \in \mathbb{R}^{h \times w}$ for $q^l$/$q^h$ simultaneously:
\begin{equation}
    S_{Gau} = \phi^G(q^l) + \phi^G(q^h), S_{Per} = \phi^P(q^l) + \phi^P(q^h).
\end{equation}
We scale above anomaly scores to [0, 1]:
\begin{equation}\
\left\{
\begin{aligned}
    S'_{Gau} &= \frac{S_{Gau}-{\min}(S_{Gau})}{{\max}(S_{Gau})-{\min}(S_{Gau})},\\
    S'_{Per} &= \frac{S_{Per}-{\min}(S_{Per})}{{\max}(S_{Per})-{\min}(S_{Per})}.\\
\end{aligned}
\right.
\end{equation}

Then the anomaly scores of a test image is acquiblue by averaging $S'_{Gau}\in \mathbb{R}^{h \times w}$ and $S'_{Per}\in \mathbb{R}^{h \times w}$:
\begin{equation}
    S' = \frac{1}{2}(S'_{Gau} + S'_{Per}).
\end{equation}
$S'\in \mathbb{R}^{h \times w}$ is interpolated to obtain the final anomaly score map $S\in \mathbb{R}^{H \times W}$. The anomaly detection score $S_A$ for each test image is determined by selecting the maximum score of $S$.

\section{Experiments}
\subsection{Experimental Setups}
\noindent
\textbf{Datasets.} 
We conduct a range of experiments on MVTec AD~\cite{bergmann2019mvtec} and VisA~\cite{zou2022spot}. MVTec AD dataset consists of a total of 15 categories and 5,354 images, with 3,629 images for training and 1,725 images for testing. The training data comprises only normal images, while the testing data includes both normal and anomaly images. VisA dataset contains 12 categories and 10,821 images, including 9,621 normal and 1,200 anomalous samples. Our method is consistent with previous FSAD methods in the use of only normal samples for training.

\noindent
\textbf{Evaluation metrics.}
For evaluating the performance of sample-level anomaly detection, we use Area Under the Receiver Operator Curve (AUROC$_i$). For anomaly localization, pixel-wise AUROC (AUROC$_p$) and Per-Region Overlap (PRO) are used as evaluation metrics.
\begin{figure}[bp]
    \centering
    \includegraphics[width=\linewidth]{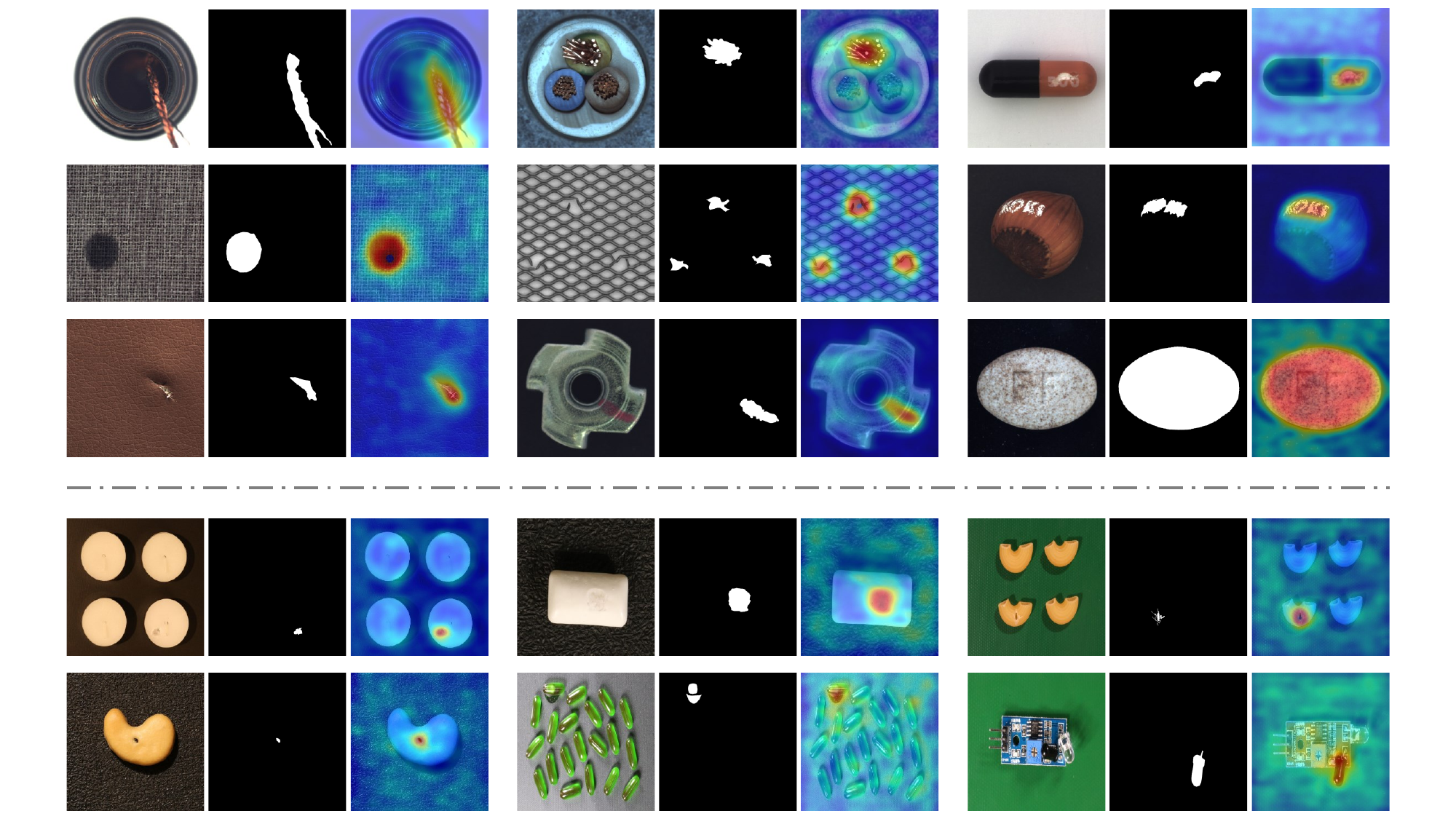}
    \caption{\textbf{Visualization results of anomaly localization} on MVTec AD dataset and VisA dataset.}
    \label{visualization}
\end{figure}

\noindent
\textbf{Implementation details.}
All experiments are implemented on an RTX 3090 GPU. 
Our experimental setup involved randomly selecting normal samples from source samples for few-shot setting and resizing all images to a resolution of $256\times256$ . For data augmentation, we generate pseudo-anomalous images as described in \cref{anomaly generation}. Specifically, a training normal image is randomly rotated within (-90, 90) degrees, and an image-level anomaly generation strategy in \cref{eq:anomaly image} is applied with a probability of 70\%. Channel-wise standardization is performed with the mean [0.485, 0.456, 0.406] and standard deviation [0.229, 0.224, 0.225]. We will obtain $N=80$ pseudo-anomalous images by above process. We adopt pre-trained models with ImageNet~\cite{deng2009imagenet} as the backbone. 
By default, WideResNet-50 is utilized as the backbone following SimpleNet~\cite{liu2023simplenet}, and features of level 2 + 3 are chosen as local features. We employ Adam optimizer~\cite{kingma2014adam}, setting the learning rate to 5e-4 for the feature adaptor, 2e-4 for the Gaussian discriminator, and 1e-4 for the Perlin discriminator. In \cref{eq:total loss}, we set $\lambda_{Per} = 2$, $\lambda_{Sim} = 0.02$ for MVTec AD~\cite{bergmann2019mvtec}, and $\lambda_{Per} = 1$, $\lambda_{Sim} = 1$ for VisA~\cite{zou2022spot}. The training is conducted over 80 epochs with a batch size of 8.

\subsection{Experimental Results}
\begin{table}[H]
  \caption{Comparison of average FSAD performance on MVTec AD and VisA dataset. \textbf{Bold} and \underline{underline} represent optimal and sub-optimal results, respectively.}
  \centering
  \resizebox{0.9\textwidth}{!}{
  \begin{tabular}{l|c|ccc|ccc|ccc}
    \toprule
    \multirow{2}{*}{Dataset} & \multirow{2}{*}{Method} & \multicolumn{3}{c|}{1-shot} & \multicolumn{3}{c|}{2-shot} & \multicolumn{3}{c}{4-shot} \\
    \cmidrule(lr){3-5}\cmidrule(lr){6-8}\cmidrule(lr){9-11}
     &&AUROC$_i$ & AUROC$_p$  & PRO & AUROC$_i$ & AUROC$_p$  & PRO& AUROC$_i$ & AUROC$_p$  & PRO\\
    \midrule
    \multirow{8}{*}{MVTec}
    &SPADE~\cite{cohen2020sub} &81.0 &91.2 &83.9 &82.9  &92.0 &85.7 &84.8 &92.7 &87.0  \\
    &PaDiM~\cite{defard2021padim} &76.6 &89.3 &73.3 &78.9 &91.3  &78.2 &80.4 &92.6 &81.3 \\
    &RegAD~\cite{huang2022registration} &- &- &- &85.7 &94.6 &- &88.2 &95.8 &- \\
    &PatchCore~\cite{roth2022towards} &83.4 &92.0 &79.7 &86.3 &93.3 &82.3 &88.8 &94.3 &84.3 \\
    &GraphCore~\cite{xie2023pushing} &89.9 &\underline{95.6} &- &91.9 &\underline{96.9} &- &92.9 &\underline{97.4} & -\\
    &WinCLIP~\cite{jeong2023winclip} &93.1 &95.2 &\underline{87.1} &94.4 &96.0 &\underline{88.4} &\underline{95.2} &96.2 &\underline{89.0} \\
    &FastRecon~\cite{fang2023fastrecon} &- &- &- &91.0 &95.9 &- &94.2 &97.0  &-\\
    &AnomalyGPT~\cite{gu2024anomalygpt} & \textbf{94.1}  & 95.3 &- &\underline{95.5}  &95.6 &- & 96.3 &96.2  &-\\
    &Ours &\underline{93.3}  & \textbf{96.2}  & \textbf{88.4}& \textbf{95.7}   & \textbf{97.2}  & \textbf{88.9} &\textbf{96.9} & \textbf{97.5} & \textbf{89.9}  \\  
    \midrule
    \multirow{5}{*}{VisA}
    &SPADE~\cite{cohen2020sub}       &79.5 &95.6 &84.1 &80.7 &96.2 &85.7 &81.7 &96.6 &87.3\\
    &PaDiM~\cite{defard2021padim}    &62.8 &89.9 &64.3 &67.4 &92.0 &70.1 &72.8 &93.2 &72.6\\
    &PatchCore~\cite{roth2022towards}&79.9 &95.4 &80.5 &81.6 &96.1 &82.6 &85.3 &96.8 &84.9\\
    &WinCLIP~\cite{jeong2023winclip} &83.8 &\underline{96.4} &85.1 &84.6 &\underline{96.8} &86.2 &\underline{87.3} &\underline{97.2} & \textbf{87.6}\\
    &AnomalyGPT~\cite{gu2024anomalygpt} & \textbf{87.4} & 96.2 &- &\textbf{88.6} & 96.4 &- & \textbf{90.6}  &96.7  &-\\
    &Ours &\underline{84.2} &\textbf{96.8} &\textbf{86.2}& \underline{87.4}  & \textbf{97.1}  & \textbf{86.3} & \underline{88.7}  &\textbf{97.2}  &86.8\\    
    \bottomrule
  \end{tabular}
  }
  \label{tab: experiments of mvtec and visa}
\end{table}
\noindent
\textbf{Few-shot anomaly detection and localization.}
We compare our DFD with prior methods specifically designed for few-shot setting.
In \cref{tab: experiments of mvtec and visa}, we illustrate average experimental results for MVTec AD~\cite{bergmann2019mvtec} and VisA~\cite{zou2022spot}. \textbf{\textit{1)} For few-shot anomaly detection}, across both datasets, our method DFD outperforms prior works. Specifically, we improve AUROC$_i$ upon the current sota FSAD approach WinClip~\cite{jeong2023winclip} by +0.2\%, +1.3\%, +1.5\% on MVTec AD and +0.4\%, +2.8\%, +1.5\% on VisA for 1, 2, 4-shot setting, respectively. \textbf{\textit{2)} For few-shot anomaly localization}, we improve AUROC$_p$ upon WinClip~\cite{jeong2023winclip} by +1.0\%, +1.2\%, +1.3\% on MVTec AD and +0.4\%, +0.3\%, +0.0\% on VisA for 1, 2, 4-shot setting. The visualization results of anomaly localization in \cref{visualization} further demonstrates the accuracy of our method in localizing anomalies.

\noindent
\textbf{Comparison with full-shot methods.}
In \cref{table: comparison with full-shot}, we compare our method with full-shot anomaly detection methods. The results show that the proposed DFD is competitive with full-shot methods. Notably, our 4-shot AUROC$_p$ surpasses that of DRAEM, which utilizes the entire set of normal samples.

\noindent \textbf{Comparison with SimpleNet.}
We choose SimpleNet~\cite{liu2023simplenet} as our baseline, a current state-of-the-art method for full-shot anomaly detection. We further conduct a range of experiments on MVTec AD~\cite{bergmann2019mvtec} and VisA~\cite{zou2022spot} datasets under few-shot settings using SimpleNet baseline. As shown in \cref{tab:comparison with simplenet}, compablue with SimpleNet~\cite{liu2023simplenet}, our proposed DFD 
\begin{wraptable}{r}{0.5\textwidth}
\vspace{-0.2cm}
  \centering
  \caption{Comparison of the flops and inference time.}
  \resizebox{1.\linewidth}{!}{
  \begin{tabular}{@{}cccc}
    \toprule
     Model & Training Time (s) $\downarrow$ & Training Flops (G) $\downarrow$& Inference Speed (s) $\downarrow$\\
    \midrule
     Ours & 10.1 & 59.3 & 0.09 \\ 
     RegAD\cite{huang2022registration} & 43.5 & 73.3 & 0.34 \\  
     \bottomrule
  \end{tabular}
  }
  \label{speed}
\vspace{-0.2cm}
\end{wraptable}
has achieved significant improvements in various metrics for FSAD.

\noindent \textbf{Effectiveness comparison with meta-learning-based methods.}
Meta-learning requires training on multiple tasks, meaning that each training step typically involves the training and evaluation of numerous subtasks. This significantly increases both computational load and memory consumption. 
In Model-Agnostic Meta-Learning (MAML)~\cite{maml}, the computation of second-order derivatives is required for each parameter update, which places a significant demand on computational resources. As shown in Tab.\ref{speed}, we compare our method with the meta-learning-based method RegAD\cite{huang2022registration} in terms of training time, training flops and inference time. The other meta-learning-based method MetaFormer\cite{wu2021learning} is not open source. The training time is the average training time for one epoch of a category. The inference speed is the average time of test time for an image.

\begin{table}
  \centering
  \caption{Comparison with full-shot methods in AUROC$_i$ and AUROC$_p$ on MVTec AD dataset.}
  \resizebox{0.55\linewidth}{!}{
  \begin{tabular}{@{\hspace{0pt}}c@{\hspace{40pt}}c@{\hspace{40pt}}c@{\hspace{40pt}}c}
    \toprule
     Model & Setting & AUROC$_i$ & AUROC$_p$ \\
    \midrule
    \multirow{3}{*}{DFD (Ours)}
     & 1-shot & 93.3 & 96.2 \\  
     & 2-shot & 95.7 & 97.2  \\  
     & 4-shot & 96.9 & 97.5 \\  
    \midrule
     SimpleNet~\cite{liu2023simplenet} & full-shot & 99.6&98.1 \\  
     PatchCore~\cite{roth2022towards} & full-shot & 99.1&98.1 \\  
     CFLOW~\cite{gudovskiy2022cflow} & full-shot & 98.3 &98.6\\  
     DRAEM~\cite{zavrtanik2021draem} & full-shot & 98.0 &97.3\\  
     \bottomrule
  \end{tabular}
  }
  \label{table: comparison with full-shot}
\end{table}

\begin{table}[hbp]
  \centering
  \caption{Comparison of average FSAD performance on MVTec AD dataset with SimpleNet. \textbf{Bold} represents optimal results.}
  \begin{tabular}{l|c|ccc|ccc}
    \toprule
     \multirow{2}{*}{Setting} & \multirow{2}{*}{Method} & \multicolumn{3}{c|}{MVTec AD} & \multicolumn{3}{c}{VisA}\\
    \cmidrule(lr){3-5}\cmidrule(lr){6-8}
    & & AUROC & p-AUROC  & PRO & AUROC & p-AUROC  & PRO\\
    \midrule
    \multirow{2}{*}{1-shot}
     & SimpleNet & 76.6 & 74.1  & 46.7 &57.1&74.0 &32.3\\
     & ours & \textbf{93.3}  & \textbf{96.2}  & \textbf{88.4} & \textbf{84.2}  & \textbf{96.8}  & \textbf{86.2}\\     
    \midrule
    \multirow{2}{*}{2-shot}
     & SimpleNet & 77.5 & 74.4  & 47.9 &62.9&80.2&38.3\\
     & ours & \textbf{95.7}  & \textbf{97.2}  & \textbf{88.9}& \textbf{87.4}  & \textbf{97.1}  & \textbf{86.3} \\     
    \midrule
    \multirow{2}{*}{4-shot}
     & SimpleNet & 78.9 & 80.8  & 56.8 &66.2&81.6&40.5 \\
     & ours & \textbf{96.9}  & \textbf{97.5}  & \textbf{89.9}& \textbf{88.7}  & \textbf{97.2}  & \textbf{86.8} \\    
    \bottomrule
  \end{tabular}
  \label{tab:comparison with simplenet}
\end{table}
\subsection{Ablation Study}
In this section, we verify the effectiveness of proposed various modules. We conduct extensive experiments on MVTec AD dataset~\cite{bergmann2019mvtec} for 2-shot setting following prior work~\cite{xie2023pushing}.

\noindent
\textbf{Influence of different components.}
We conduct the following experiments: \textit{\textbf{(1)}} Baseline (SimpleNet~\cite{liu2023simplenet}, i.e. Gaussian Discriminator and pseudo-anomalies at feature-level), denoted as Gaussian-Disc; \textit{\textbf{(2)}} Adding Perlin Discriminator and pseudo-anomalies at image-level, denoted as Perlin-Disc ; \textit{\textbf{(3)}} Adding both Perlin-Disc and data augmentation (DA); \textit{\textbf{(4)}} Adding multi-frequency information construction (MFIC) module to (3); \textit{\textbf{(5)}} Adding similarity loss ($\mathcal{L}_{Sim}$) to (3); \textit{\textbf{(6)}} Proposed DFD without Perlin-Disc; 
\textit{\textbf{(7)}} Proposed DFD without Gaussian-Disc;
\textit{\textbf{(8)}} Proposed DFD in this paper.
As shown in \cref{table: influence of different modules}, our baseline (SimpleNet~\cite{liu2023simplenet}) only obtains 77.5\%/74.4\% AUROC$_i$/AUROC$_p$ because of its poor utilization of a limited number of normal images. Training with our Perlin Discriminator can increase the AUROC$_i$/AUROC$_p$ by +2.1\%/+10.9\%. When we add DA \textbf{into} above modules, the performance increases by +12.0\%/+9.8\%. Subsequently, adding MFIC module improves by +2.7\%/+1.2\%. Introducing similarity loss ($\mathcal{L}_{Sim}$) can enhance performance by an additional +2.0\%/+0.9\%. The other loss functions are specifically tailored to guide the training of their respective discriminators, thus obviating the need for additional experimental validation of their efficacy. \cref{table: influence of different modules} shows that each module added improves model performance.
\begin{table}[ht]
  \centering
  \caption{Performance with the configuration of different components.}
  \resizebox{0.6\linewidth}{!}{
  \begin{tabular}{@{\hspace{0pt}}c@{\hspace{10pt}}c@{\hspace{10pt}}c@{\hspace{10pt}}c@{\hspace{10pt}}c@{\hspace{10pt}}c}
    \toprule
    Gaussian-Disc &  Perlin-Disc & DA & MFIC & $\mathcal{L}_{Sim}$ & Performance\\
    \midrule
    \cmark & \xmarkg & \xmarkg & \xmarkg & \xmarkg & 77.5/74.4/47.9\\
    \cmark & \cmark & \xmarkg & \xmarkg & \xmarkg & 79.6/85.3/61.3\\
    \cmark & \cmark & \cmark & \xmarkg & \xmarkg & 91.6/95.1/84.6 \\
    \cmark & \cmark & \cmark & \cmark & \xmarkg & 93.7/96.3/88.9 \\
    \cmark & \cmark & \cmark & \xmarkg & \cmark & 93.1/96.5/88.0 \\
    \cmark & \xmarkg & \cmark & \cmark & \cmark & 92.9/96.4/86.2 \\
    \xmarkg & \cmark & \cmark & \cmark & \cmark & 94.0/93.4/83.7 \\
    \cmark & \cmark & \cmark & \cmark & \cmark & 95.7/97.2/88.9\\
    \bottomrule
  \end{tabular}
  }
  \label{table: influence of different modules}
\end{table}

\noindent
\textbf{Influence of dual-path discriminators and different structures of Perlin Discriminator.}
We run separate experiments using different discriminators with the results in rows 6 and 7 of \cref{table: influence of different modules}. The performance of using a single discriminator individually deteriorated in comparison to using dual-path discriminators.

We experiment with different structures of Perlin Discriminator, and the results illustrates our improvements, achieving +1.2\% AUROC$_i$ and +1.2\% AUROC$_p$ over 2-layer MLP in \cref{table:different structure of Perlin Discriminator}. The 2-layer MLP is the same as the Gaussian Discriminator following SimpleNet~\cite{liu2023simplenet}. We believe that the translation invariance of ViT makes it sensitive to positional information. What's more, ViT can capture global context information, which allows it to locate anomalies more accurately for generated pseudo-anomalies at the image level.
\begin{table}[hb]
  \centering
  \caption{Ablation of different structures for Perlin Discriminator.}
  \begin{tabular}{@{\hspace{15pt}}l@{\hspace{20pt}}c@{\hspace{20pt}}c@{\hspace{20pt}}c}
    \toprule
     Perlin Discriminator & AUROC$_i$& AUROC$_p$ &PRO \\
    \midrule
     A single-MLP + a single-ViT  (Ours) & \textbf{95.7} & \textbf{97.2}&\textbf{88.9} \\  
     2-layer MLP  & 94.5 & 96.0&88.8\\  
     \bottomrule
  \end{tabular}
  \label{table:different structure of Perlin Discriminator}
\end{table}

\noindent
\textbf{Influence of different frequency information.}
Different frequency components of an image represent different information. As shown in \cref{table: ablation frequency}, we conduct a series of experiments to investigate the impact of using different frequency components: (1) the proposed DFD; (2) without multi-frequency information construction; (3) only high-frequency information; (4) only low-frequency information. The results indicate that using only high-frequency information demonstrates superior performance compared to using only low-frequency information. Using the original image performs better than using high-/low-frequency information alone. However, incorporating high-frequency and low-frequency information performs the best, suggesting the normal images and abnormal images contain complementary frequency information.

\begin{figure}
    \centering
    \includegraphics[width=\linewidth]{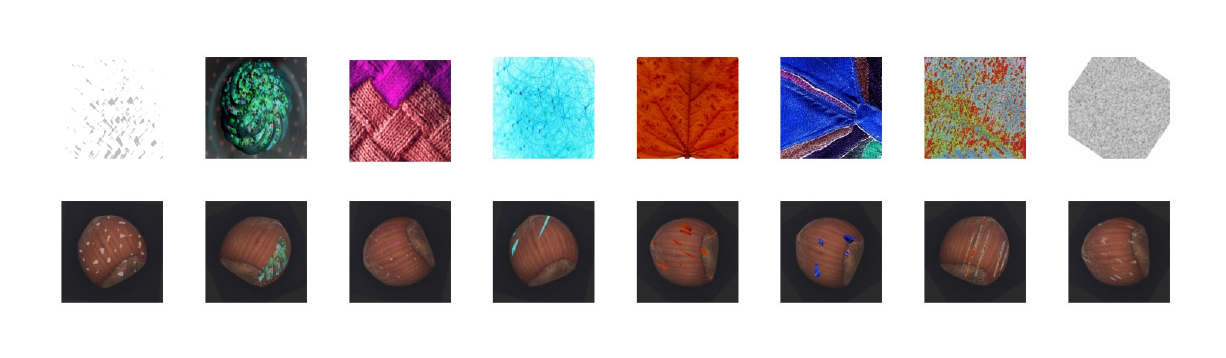}
    \caption{Augmented image-level pseudo-anomalous images. The line above represents texture images from the DTD dataset\cite{cimpoi2014describing}, and the line below represents the image-level pseudo-anomaly images. }
    \label{pseudo-anomalies}
\end{figure}

\noindent
\textbf{Influence of different forms of anomalies.}
The introduction of pseudo-anomalies at both the image and feature levels exerts significant influence for our dual-path discriminators to learn a joint representation of anomalous features and normal features. As illustrated in \cref{table: ablation anomalies}, omitting any specific type of pseudo-anomaly results in a deterioration of performance. "I-anomaly"/"F-anomaly" denotes that only image-level/feature-level pseudo-anomaly is used in experimental setting and "W/o anomaly" indicates that no anomaly generation is performed during training. The \cref{pseudo-anomalies} shows some examples of image-level anomalies. The feature representations learned by the discriminators during training fails to grasp the intricacies of the anomalies. The malfunctioning of any of the discriminators can negatively impact the overall performance, resulting in a decline in the final results.

\begin{table}[hb]
\centering
\begin{minipage}[t]{0.45\linewidth}
  \centering
  \caption{Ablation study of different frequency information.}
  \resizebox{0.98\linewidth}{!}{
  \begin{tabular}{@{}cccc}
    \toprule
     Model & AUROC$_i$& AUROC$_p$ & PRO \\
    \midrule
     Ours & \textbf{95.7} & \textbf{97.2}&\textbf{88.9} \\  
     W/o MFIC & 93.3 & 96.2&70.3 \\  
     High-frequency & 92.5 & 94.2&80.4\\
     Low-frequency & 91.7  & 94.1& 73.6\\
     \bottomrule
  \end{tabular}
  }
  \label{table: ablation frequency}
\end{minipage}
\begin{minipage}[t]{0.45\linewidth}
  \centering
  \caption{Ablation study of different forms of anomalies.}
  \resizebox{0.98\linewidth}{!}{
  \begin{tabular}{@{}cccc}
    \toprule
     Model & AUROC$_i$& AUROC$_p$ & PRO \\
    \midrule
     Ours & \textbf{95.7} & \textbf{97.2}&\textbf{88.9} \\ 
     W/o anomaly & 59.7 & 36.3&\phantom{8}8.9 \\  
     I-anomaly& 91.3 & 87.4&39.1\\
     F-anomaly & 83.7  & 91.4& 70.7\\
     \bottomrule
  \end{tabular}
  }
  \label{table: ablation anomalies}
\end{minipage}
\end{table}

\begin{figure}[!h]
    \centering
    \includegraphics[width=0.8\linewidth,trim=100 370 100 0,clip]{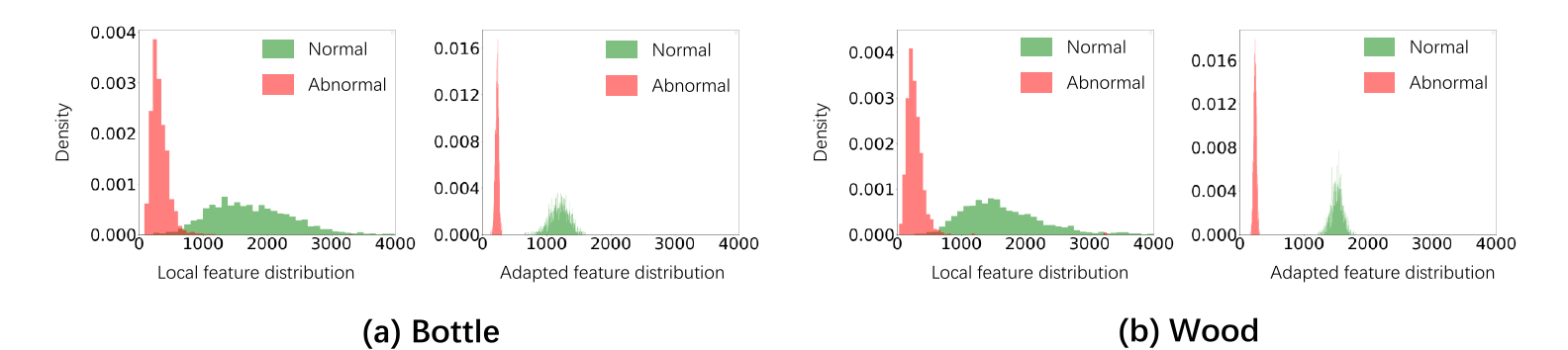}
    \caption{\textbf{Log-likelihood histograms} from bottle and wood category. Left is local feature without adaptor, right is the adapted features with adaptor.}
    \label{fig:ablation: adaptor}
\end{figure}

\noindent
\textbf{Influence of feature adaptor.}
The pre-trained backbone utilizes the ImageNet~\cite{deng2009imagenet} for training, which significantly differs from industrial images. To reduce domain bias by these different distributions, we use a feature adaptor in fine-grained feature construction module. In \cref{fig:ablation: adaptor} the features with a feature adaptor becomes more compact and the boundary between normal and abnormal distributions becomes clearer. Moreover, different from SimpleNet~\cite{liu2023simplenet}, a similarity loss $\mathcal{L}_{Sim}$ is expected to push normal features apart from normal features. In \cref{table: influence of different modules}, quantitative results also illustrate our similarity loss $\mathcal{L}_{Sim}$ enhances AD performance.

\noindent
\textbf{Influence of loss function.}
We compare the commonly used classification loss function and the proposed truncated $l_1$ loss. Specifically, we replace the truncated $l_1$ loss in Gaussian loss $\mathcal{L}_{Gau}$ (\cref{gaussian loss}) and pixel loss (\cref{pixel loss}), with cross-entropy loss, focal loss, and MSE loss,
denoted as "Ours-CE", "Ours-Focal", and "Ours-MSE" respectively. The
results depicted in \cref{table: loss function} clearly demonstrate that our truncated $l_1$ loss yields the most favorable outcomes.
\begin{wraptable}{r}{0.5\textwidth}
  \centering
  \caption{Ablation study of different loss function.}
  \resizebox{1.\linewidth}{!}{
  \begin{tabular}{@{\hspace{5pt}}l@{\hspace{40pt}}c@{\hspace{40pt}}c@{\hspace{40pt}}c}
    \toprule
     Model & AUROC$_i$& AUROC$_p$ &PRO \\
    \midrule
     Ours & \textbf{95.7} & \textbf{97.2}&\textbf{88.9} \\
     Ours-CE  & 94.0 & 96.8&84.0\\  
     Ours-Focal  & 94.8 & 96.3&82.0\\  
     Ours-MSE  & 93.6 & 96.2&88.6\\  
     \bottomrule
  \end{tabular}
  }
  \label{table: loss function}
  \vspace{-0.2cm}
\end{wraptable}

\begin{figure}[!ht]
    \centering
    \includegraphics[width=0.5\linewidth,trim=270 420 230 0,clip]{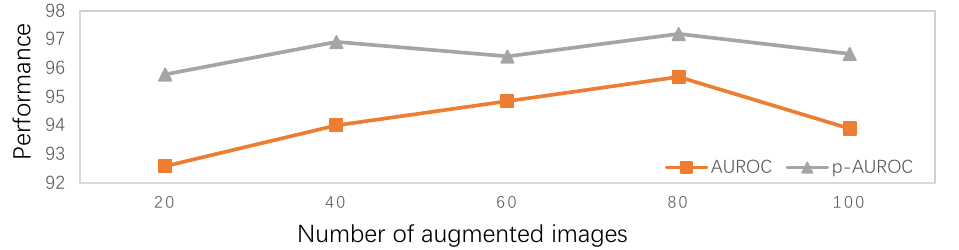}
    \caption{\textbf{Performance of the number of augmented images} per training normal image.}
    \label{fig:num}
\end{figure}

\noindent
\textbf{Influence of the number of augmented images.}
We primarily enhance the utilization rate of samples through data augmentation. We investigate the influence of the number of augmented images per normal image. As shown in \cref{fig:num}, within a certain range, an increased quantity of augmented images correlates positively with enhanced performance. However, when the number becomes excessively large, the performance may deteriorate. Thus we choose the number of augmented images to be $N=80$. 

\section{Conclusion}
\noindent\textbf{Conclusion.} In this paper, we propose a novel and simple DFD approach from a frequency perspective for few-shot anomaly detection, addressing a significant issue in industrial smart manufacturing. We generate anomalies at both image-level and feature-level to fully utilize the limited number of source normal images. To better train the feature adaptor, we introduce a similarity loss to push normal features apart from abnormal features. We further employ dual-path discriminators to estimate abnormality for two different forms of anomalies. In the end, our DFD network is capable of learning a joint representation of the features of both normal and abnormal images.

\noindent\textbf{Limitation.} Although our method DFD exhibits favorable performance, the generated pseudo-anomalies at image-level and feature-level still differ from real anomalies on industrial images. Additionally,  data augmentation for each source normal images would increase the training time and may lead to model over-fitting.
\section*{Acknowledgment}
This work is supported  in part by the National Natural Science Foundation of China under Grant 62303405, in part by Ningbo Natural Science Foundation project under Grant 2023J400, and in part by Ningbo Key Research and Development Plan under Grant 2023Z116.
\bibliography{main}
\bibliographystyle{elsarticle-num}

\end{document}